\documentclass[5p,times]{elsarticle}
\usepackage[utf8]{inputenc}
\usepackage{amsmath}
\usepackage{xcolor}
\usepackage{graphicx}
\usepackage{url}
\usepackage[ruled,vlined]{algorithm2e}
\usepackage{placeins}

\SetKwInput{KwInput}{Input}
\SetKwInput{KwOutput}{Output}

\begin{document}

\begin{frontmatter}

\title{A scalable algorithm for the optimization \\ of neural network architectures}

\author[rvt]{Massimiliano Lupo Pasini\corref{cor1}} \ead{lupopasinim@ornl.gov}
\author[rvt2]{Junqi Yin}
\author[rvt3]{Ying Wai Li}
\author[rvt2]{Markus Eisenbach}

\address[rvt]{Oak Ridge National Laboratory, Computational Sciences and Engineering Division, 1 Bethel Valley Road, Oak Ridge, TN, USA, 37831}
\address[rvt2]{Oak Ridge National Laboratory, National Center for Computational Sciences, 1 Bethel Valley Road, Oak Ridge, TN, USA, 37831}
\address[rvt3]{Los Alamos National Laboratory, Computer, Computational, and Statistical Sciences Division, Los Alamos, NM, 87545, USA}

\cortext[cor1]{Corresponding author}

\date{}

\begin{abstract}
    We propose a new scalable method to optimize the architecture of an artificial neural network. The proposed algorithm, called Greedy Search for Neural Network Architecture, aims to determine a neural network with minimal number of layers that is at least as performant as neural networks of the same structure identified by other hyperparameter search algorithms in terms of accuracy and computational cost.
    Numerical results performed on benchmark datasets show that, for these datasets, our method outperforms state-of-the-art hyperparameter optimization algorithms in terms of attainable predictive performance by the selected neural network architecture, and time-to-solution for the hyperparameter optimization to complete. 
\end{abstract}


\end{frontmatter}

{\footnotesize \noindent This manuscript has been authored in part by UT-Battelle, LLC, under contract DE-AC05-00OR22725 with the US Department of Energy (DOE). The US government retains and the publisher, by accepting the article for publication, acknowledges that the US government retains a nonexclusive, paid-up, irrevocable, worldwide license to publish or reproduce the published form of this manuscript, or allow others to do so, for US government purposes. DOE will provide public access to these results of federally sponsored research in accordance with the DOE Public Access Plan (\url{http://energy.gov/downloads/doe-public-access-plan}).}

\normalsize

\section*{Introduction}

Deep neural networks (NN) are nonlinear models used to approximate unknown functions based on observational data \cite{Minsky, Neumann, Rosenblatt, Rosenblatt2}. Their broad applicability is derived from their complex structure, which allows these techniques to reconstruct complex relations between quantities selected as inputs and outputs of the model \cite{Haykin}. 
From a mathematical perspective, a NN is a directed acyclic graph where the nodes (also called neurons) are organized in layers. The type of connectivity between different layers is essential for the NN to model complex dynamics between inputs and outputs. The structure or architecture of the graph is mainly summarized by the number of layers in the graph, the number of nodes at each layer and the connectivity between nodes of adjacent layers. 

The performance of a NN is very sensitive to the choice of the architecture for multiple reasons. Firstly, the architecture strongly impacts the prediction computed by a NN. Indeed, NN's with different structures may produce different outputs for the same input. On the one hand, structures that are too simple may not be articulate enough to reproduce complex relations. This may result in underfitting the data with high bias and low variance in the predictions. On the other hand, architectures that are too complex may cause numerical artifacts such as overfitting, leading to predictions with low bias and high variance. Secondly, the topology of a NN affects the computational complexity of the model, because an increase in layers and nodes leads to an increase in floating point operations to train the model and to make predictions. 
Therefore, identifying an appropriate architecture is an important step that can heavily impact the computational complexity to train a deep learning (DL) model and the final attainable predictive power of the DL model itself. However, the parameter space of NN architectures is too large for an exhaustive search. {\it In fact, the number of architectures grows exponentially with the number of layers, the number of neurons per layer and the connections between layers}. 

Several approaches have been proposed in the literature for hyperparameter optimization (HPO) \cite{baker, Bergstra, reinforcement_NN, fahlman, Goodfellow, zemel, gupta, progressive_NN, luo, Snoek, zoph} with the goal to identify a NN architecture that outperforms the others in terms of accuracy. 
Sequential Model-Based Optimization (SMBO) algorithms \cite{Bergstra} are a category of HPO algorithm. Examples of SMBO algorithms are Bayesian Optimization (BO) \cite{Snoek, Snoek2} and its less expensive variant Tree-Parzen estimator (TPE), which rely on information available from previously trained models to guide the choice of models to build and train in following steps. The use of past information generally benefits the reduction of the number of neural networks to train in the next iterations, and provides an assessment of uncertainty by incorporating the effect of data scarcity. The efficacy of the results obtained with BO is highly sensitive to the choice of the prior distribution on the hyperparameter space as well as the acquisition function to select new points to evaluate in the hyperparameter space.   
Another class of HPO methods is represented by genetic algorithms \cite{Ettaouil, gupta2, Holland, kitano, koehn, tsai} and evolutionary algorithms (EA) \cite{menndl_paper, menndl_url}, which evolve the topology of a NN by alternatively adding or dropping nodes and connections based on results attained by previous NN models.  
Incremental, adaptive approaches \cite{Treadgold} and pruning algorithms \cite{pruning, compression} or random dropout \cite{srivastava} can also be computationally convenient because they tend to minimize the number of NN models built and trained.
All the SMBO, EA and incremental approaches described above adopt theoretical expedients \cite{domhan, hinz} to reduce the uncertainty of the hyperparameter estimate, but this comes at the price of not being scalable.

Several scalable algorithms for hyperparameter search have been proposed in the literature. Grid Search (GS), or parameter sweep, searches exhaustively through a specified subset of hyperparameters. The subset of hyperparameters and the bounds in the search space are specified manually. Moreover, the search for continuous hyperparameters requires a manually prescribed discretization policy. Although this technique is straightforwardly parallelizable, it becomes more and more prohibitive in terms of computational time and resources when the number of hyperparameters increases. Random Search (RS) \cite{random_search} differs from GS mainly in that it explores hyperparameters stochastically instead of exhaustively.
RS is likely to outperform GS in terms of time-to-solution \cite{random_search, Lia}, especially when only a small number of hyperparameters affects the final predictive power of DL model. The independence of the hyperparameter settings used by GS and RS make these approaches appealing in terms of parallelization and obtainable scalability. 
However, both GS and RS require expensive computations to perform the hyperparameter search. 

{\it We present a scalable method to determine, within a given computational budget, the NN with minimal number of layers that performs at least as well, in terms of accuracy and time-to-solution, as NN models of the same structure identified by other hyperparameter search algorithms.} { The computational budget is an important aspect of the NN training for two important reasons: the available computational power and the period of time when the computational power is available. The former imposes obvious intrinsic limitations, the latter becomes important when critical decisions have to be made in a timely and accurate manner. }
We refer to our method as \textit{Greedy Search for NN Architecture (GSNNA)}. {Although our algorithm increments the number of hidden layers adaptively, it differs from other incremental, adaptive algorithms proposed in the literature \cite{Cortes, Kwok, Liu, Friedman} in that our algorithm performs a {\it stratified} (sliced) RS restricted to one hidden layer at each iteration. { This stratified RS is the most important difference between GSNNA and previous methods.}
The selection of the NN models is driven by the validation score, which is used as a metric to quantify the predictive performance of the DL models.}
Starting with the first layer, a random search is performed in parallel on various instantiations of the DL model, to determine the optimal number of neurons on each layer and the hyperparameters of the associated DL model.  
Random search would identify the hyperparameters for each of the instantiations, and the performance of the DL model would determine the best number of neurons and retain the hyperparameters associated with best performing model. The same sliced RS procedure is applied to the next layers.
 The recycling of information from previously evaluated models guarantees a fine level of \textit{exploitation}, and the stratified RS performed at each iteration still guarantees a thorough (albeit not exhaustive) \textit{exploration} of the objective function landscape in the hyperparameter space to prevent stagnations at local minima. \textit{By performing a stratified RS at each iteration, our new approach retains a high level of parallelization, because the NN models can be trained concurrently at each step.}

In this work we focus on two widely used NN architectures: multi-layer perceptrons (MLP) and convolutional NN models (CNN). The performance of the HPO algorithms is evaluated using five standard datasets, each of them is associated with its specifically tailored DL model. The validation of the method will be done by comparing the efficiency of the DL model on the determined NN architecture with the efficiency of the same type of NN identified by other algorithms.

The paper is organized in five sections. Section \ref{NN_section} introduces the DL background.
Section \ref{adaptive_algorithm} explains our novel optimization algorithm for the architecture of NN models. Section \ref{algorithm_implementation} describes the computational environment where the numerical experiments are performed, the benchmark datasets, the specifics of the implementations for the each HPO algorithm considered, and the parameter setting for each HPO algorithm. Section \ref{numerical_results} presents numerical experiments where we compare the performance of our HPO algorithm with Bayesian Optimization and Tree-Parzen Estimator.
Section \ref{conclusions} summarizes the results presented and describes future directions to possibly pursue.

\section{Deep learning background}
\label{NN_section}
Given an unknown function $f$ that relates inputs $x$ and outputs $y$ as follows
\begin{equation}
    y = f(x),
    \label{function}
\end{equation}
a \textit{deep feedforward network}, also called \textit{feedforward neural network} or \textit{multilayer perceptron} (MLP) \cite{Haykin, Goodfellow}, is a predictive statistical model that approximates the function $f$ by composing together many different functions such that 
\begin{equation}
\hat{f}(\mathbf{x}) = f_{L+1}(\cdots f_{\ell+1}(f_\ell(f_{\ell-1}(\ldots f_0(\mathbf{x}))))),
\label{composition}
\end{equation}
{ where $\hat{f}:\mathbb{R}^p\rightarrow \mathbb{R}^b$, and $f_\ell:\mathbb{R}^{p_{\ell}}\rightarrow \mathbb{R}^{p_{\ell+1}}$ for $\ell=0,\ldots,L+1$.}
The goal is to identify the proper number $L$ so that the composition in Equation \eqref{composition} resembles the unknown function $f$ in \eqref{function}. The composition in Equation \eqref{composition} is modeled via a directed acyclic graph describing how the functions are composed together. 
The number $L$ that quantifies the complexity of the composition is equal to the number of hidden layers in the NN.
We refer to the input layer as the layer with index $\ell=0$. The indexing for hidden layers of the deep NN models starts with $\ell=1$. In this section we consider a NN with a total of $L$ hidden layers. The symbol $p_\ell$ is used to denote the number of neurons at the $\ell$th hidden layer. Therefore, $p_0$ coincides with the dimensionality of the input, that is $p_0 = p$. The very last layer with index $L+1$ represents the output layer, meaning that $p_{L+1}=b$ coincides with the dimensionality of the output. We refer to $\mathbf{w}\in \mathbb{R}^{N_{tot}}$ as the total number of regression coefficients. Following this notation, the function $f_0$ corresponds to the first layer of the NN, $f_1$ is the second layer (first hidden layer) up to { $f_{L+1}$} that represents the last layer (output layer).  
In other words, deep feedforward networks are nonlinear regression models and the { non-linearity} is given by the composition in Equation \eqref{composition} to describe the relation between predictors $\mathbf{x}$ and targets $\mathbf{y}$. 
This approach can be reinterpreted as searching for a mapping that minimizes the discrepancy between values $\hat{\mathbf{y}}$ predicted by the model and given observations $\mathbf{y}$. 

Given a dataset with $m$ data points, the process of predicting the outputs for given inputs via an MLP can thus be formulated as 
\begin{equation}
\hat{\mathbf{y}} = F(\mathbf{x}, \mathbf{w}),
\end{equation}
where the operator $F:\mathbb{R}^{p_0}\times \mathbb{R}^{N_{tot}}\rightarrow \mathbb{R}^b$ is
\begin{equation}
\begin{aligned}
    F(\mathbf{x}, \mathbf{w}) & = \varphi_{L+1}\bigg( \sum_{k_{L}}w_{k_{L+1} k_{L}}\varphi_L \bigg( \sum_{k_{L-1}} w_{k_L k_{L-1}}\varphi_{L-1}\bigg(\ldots \\& \ldots \varphi_1\bigg( \sum_{i=1}^{}w_{k_1i}x_i \bigg) \bigg) \bigg) \bigg),
    \label{F_entry}
\end{aligned}
\end{equation}
{ where $\varphi_{\ell}$ ($\ell=1,\ldots,L+1$) are activation functions used to generate non-linearity in the predictive model.}
Using the matrix notation for the weights connecting adjacent layers as
\begin{equation}
W_{\ell,\ell-1} \in \mathbb{R}^{p_\ell \times p_{\ell-1}},
\end{equation}
we can rewrite \eqref{F_entry} as
\begin{equation}
F(\mathbf{x}, \mathbf{w}) = \varphi_{L+1}\bigg(W_{L+1, L}\bigg(\varphi_{L}\bigg( \ldots \bigg( \varphi_1\bigg (W_{1,0}\mathbf{x}\bigg ) \bigg) \bigg) \bigg )\bigg).
\label{F_matrix}
\end{equation}
{ The composition of the activation functions $\varphi_{\ell}$ with the tensor products using matrices $W_{\ell+1,\ell}$ at the $\ell$th layer corresponds to the $f_{\ell}$ used in Equation \eqref{composition}. }
The notation in \eqref{F_matrix} highlights that $N_{tot}$ is the total number of regression weights used by the NN. This value must account for all the entries in $W_{\ell, \ell-1}$'s matrices, that is 
\begin{equation}
N_{tot} = \sum_{\ell=1}^{L+1} p_\ell p_{\ell-1}.
\end{equation}
If the target values are continuous quantities, the very last layer $\varphi_{L+1}$ is usually chosen to be linear, i.e., the identity function. If the target values are categorical, then $\varphi_{L+1}$ is usually set to be the logit function. 
If the number of hidden layers is set to $L=0$ and $\varphi_1$ is set to be the identity function, then the statistical model becomes a classical linear regression model. If the number of hidden layers is set to $L=0$ and $\varphi_1$ is set to be the logit function, then the statistical model becomes a logistic regression model. 

In order to exploit local correlations in the data, convolutional kernels can be composed with the activation functions $\varphi_i$.
Convolution is a powerful mathematical tool that models local interactions between data points. {As such, convolution uses the same set of regression coefficients to model local interactions across the entire data instead of using several sets of regression coefficients, one specific for each neighbourhood as a standard MLP architecture would require. The use of the convolution thus significantly reduces the dimensionality of the coefficients needed in DL models to reconstruct local features in regularly structured data.} Well known examples of data that respect this geometrical properties are images. NN models that exploit the data locality for the feature extraction are called \textit{Convolutional Neural Networks} (CNN) \cite{Fukushima, cnnpaper, LeCunn} and they are characterized by a sparse connectivity or sparse weights that stems from the sparse interaction between data. In essence CNN are the nonlinear generalization of kernel regression and they inherit from the linear case the advantages of replacing dense matrix multiplication with sparse matrix multiplications. This benefits the computation by reducing the number of FLOPS required to perform matrix multiplications, and reduces the memory requirement to store the regression weights.

\section{Adaptive selection of the number of hidden layers}
\label{adaptive_algorithm}

The goal of our novel HPO algorithm is to determine, within a given computational budget, the NN with minimal number of layers that performs at least as well on training datasets, in terms of accuracy and time-to-solution, as NN models of the same structure identified by other hyperparameter search algorithms. 
The HPO is performed over a set of hyperparameters which differs according to the type of NN architecture considered. For MLP models, the HPO is performed over the number of hidden layers, the number of neurons per layer, the type of nonlinear activation function at each hidden layer and the batch size used to train the model with a first-order optimization algorithm. 
For CNN models, the number of neurons is replaced by the number of channels. In addition, the convolutional kernel, the dropout rate and the pooling { are} optimized as well.
In order for the HPO procedure to be applied, the region of the hyperspace explored must be bounded to guarantee that the exploration is restrained within a computational budget for the number of layers and the other NN hyperparameters. 

The result of the procedure is dataset dependent, in that it aims to identify a customized neural network architecture that well predicts the input-output relation for the dataset at hand. The dataset is split into a training set, a validation set and a test set. The training portion is used to train the instantiated NN models. The performance of the DL models over the validation set is used to associate the model with a score, which is used to compare the performance of the NN instantiated. The test set is used to quantify the predictive performance of the finally selected NN model by computing the test score. We refer to Section \ref{section_comparison} for details about the metrics used to measure the performance of a NN.

\begin{figure}[thbp]
   \centering
\includegraphics[width=0.4\textwidth]{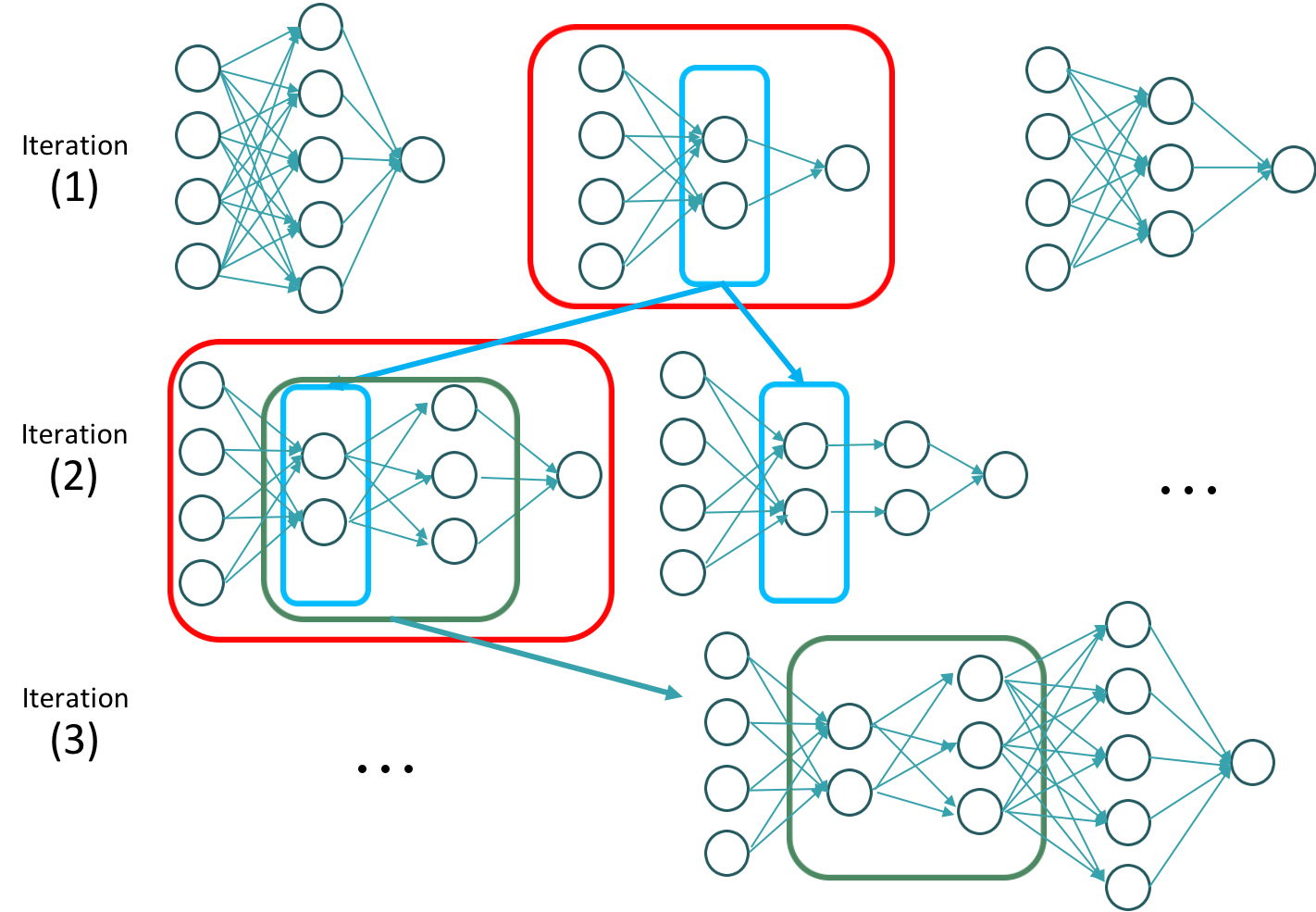}
    \caption{Illustration of the Greedy Search for Neural Network Architecture (GSNNA). The illustration explains how the architecture of the NN is enriched at each iteration. The NN models built at iteration (1) have only one hidden layer and the number of neurons inside the hidden layers is chosen via RS. Every NN is trained and the predictive performance over the validation set is measured. The NN with the best validation score is selected (circled in red). If the attained accuracy meets the requirements prescribed by the user, the algorithms stops and returns the selected NN. Otherwise, the hyperparameters of the first hidden layer are transferred to iteration (2). The NN models built at iteration (2) have the same number of neurons in the first hidden layers as the best NN from iteration (1), whereas the number of neurons at the second hidden layer is chosen with another stratified RS. The NN models are trained and the validation scores from each NN are collected. The NN with the best predictive performance is chosen (circled in red). If the performance meets the requirements, the algorithms stops and returns the selected NN. Otherwise, the information about the numbers of neurons in the first and second hidden layers are transferred to iteration (3), so that another stratified RS takes place on the number of neurons inside the third hidden layer.}
    \label{fig:fig4} 
\end{figure}

The pseudo-code that describes GSNNA is presented below, in Algorithm \ref{asnna}. 
The method starts by performing RS over NN models with one hidden layer and it selects the NN that attains the best predictive performance over the validation portion of the dataset. 
The random search identifies the hyperparameters for each of the instantiations and the performance of the deep learning model determines the best number of neurons and the hyperparameters associated with best performing model on the respective datasets to retain. 
The procedure continues by freezing the number of neurons and the hyperparameters in the previous hidden layers every time a new hidden layer is added, and the sliced RS is performed only on the hyperparameters of the last hidden layer in the architecture. This iterative procedure proceeds until either the validation score reaches a prescribed threshold or the maximum number of hidden layers is reached.
An illustration that explains how GSNNA proceeds is shown in Figure \ref{fig:fig4}. The number of neurons needed may vary from layer to layer in order for a NN architecture to attain a desired accuracy. It is thus possible that the NN may have to alternatively expand and contract across the hidden layers to properly model the nonlinear relations between input and output data. GSNNA allows this, as the number of neurons at each selected through a stratified RS may vary for each hidden layer. 

\begin{algorithm}[htb]
\SetAlgoLined
\KwInput{
\begin{itemize}
   \item $L$ = maximum number of hidden layers
   \item $N_{max\_nodes}$ = maximum number of nodes (neurons) per layer
   \item $score_{\text{threshold}}$ = threshold on the final performance prescribed
   \item $model\_eval\_iter$  = number of model evaluations per iteration
\end{itemize} 
}
\KwOutput{best\_model}
Set number of hidden layers $\ell= 1$\;
Set \text{best\_model} as linear regression (for regression problems) or logistic regression (for classification problems)\;
Compute $score$\;
\While{$score<score_{\text{threshold}}\And \ell \le L$}
{
Build $model\_eval\_iter$ NN models with $\ell$ hidden layers each\;
Set number of nodes and activation functions for first $(\ell-1)$ hidden layers as in best\_model\;
Perform random search for number of nodes in the last hidden layer and for the remaining hyper-parameters\;
Select best\_model as the NN with best performance\;
Retrieve best\_model and store info about number of nodes and activation functions per layer\;
$\ell = \ell+1$\;
}
\Return{best\_model}
\label{asnna}
\caption{Greedy Search for Neural Network Architecture (GSNNA)}
\end{algorithm}

{ The stratified RS is the most important difference between GSNNA and previous methods. The main contrast of GSNNA with respect to previous methods is the greedy approach adopted in increasing the number of hidden layers. As more hidden layers are added to the NN architecture, the predictive power of the model increases, but with this also the computational cost for training. Previous methods treat the number of hidden layers as any other hyperparameter, and the methods sometimes construct expensive neural networks at intermediate steps, and these NN are later discarded in favor of smaller ones. By performing a greedy approach on the number of hidden layers, GSNNA avoids this type of extreme situations where very expensive NN are trained and discarded through intermediate steps, and this favors a lower computational cost per iteration.} The validation of the method will be shown by comparing the efficiency of the DL model on the determined NN architecture with the efficiency of the same type of DL model identified by other algorithms. 

\subsection{Reduction of dimensionality in the hyperparameter search}
Transferring information from smaller to bigger NN models across successive iterations and restricting the RS to the hyperparameters associated only with the last hidden layers reduces the dimension of the hyperparameter space to explore. 
In this section we compare the dimensionality (number of elements in a set) of the hyperparameter space explored by a standard HPO algorithm (e.g. GS, RS, SMBO, EA) with the dimensionality of the hyperparameter space explored by GSNNA. 

Denote the maximum number of neuron per layer with \newline $N_{max\_nodes}$ and the maximum number of hidden layers with $L$. 
The number of hidden layers and the number of neurons per layer are hyperparameters that affect the structure of the NN models, whereas all the other hyperparameters affect the training of the DL model. Because GSNNA differ from state-of-the-art HPO algorithms by the way the number of hidden layers are optimized, this is the only factor that determines a change in dimensionality of the hyperparameter space. 
The stratified RS in GSNNA allows us to avoid the \textit{curse of dimensionality}, because the number of NN architectures to span at each iterations decreases from $N_{max\_nodes}^L$ (as it is for a standard HPO algorithm) to $N_{max\_nodes}$. 
The reduced dimensionality of the hyperparameter space leads also to a reduction of the uncertainty over the estimated attainable predictive performance. This is shown in the numerical experiments in Section \ref{numerical_results}, where the accuracy attained by the NN models selected from multiple runs of GSNNA has narrower confidence intervals than the ones obtained with BO and TPE, indicating that the estimates obtained with GSNNA are more reliable.  

\subsection{Computational complexity of GSNNA}
{ Let us refer to $C$ as the number of independent model evaluations performed in one iteration of an HPO algorithm, and $L$ the number of HPO iterations performed. The computational complexity of one iteration of GSNNA is $O(C)$ (and hence $O(CL)$ for the whole algorithm), because the algorithm compares the predictive performance of $C$ models and selects the best one to proceed to the next iteration. To put this value in perspective, we remind the reader that the computational complexity of one iteration of BO is cubic both in the number of independent model evaluations and in the number of iterations performed, that is $\mathcal{O}((CL)^3)$, and the computational complexity of one iteration of TPE is cubic only in the number of independent model evaluations, that is $\mathcal{O}(C^3)$. In terms of computational complexity, GSNNA thus provides a significant improvement with respect to BO and TPE, because the computational complexity per iteration is constant with respect to the iteration count, and the computational complexity of one iteration of HPO is reduced from cubic to linear. This benefit makes GSNNA appealing for scaling purposes with large values of independent model evaluations $C$. We also remind the reader that the independent model evaluations in each iteration can be performed concurrently, as we did in the numerical experiments described in Section \ref{numerical_results} of this work.} 

\section{Algorithm implementation}
\label{algorithm_implementation}
In this section we describe the computational environment where the numerical experiments were performed, the specifics of the implementations for each of the HPO algorithms considered, the benchmark datasets used and the parameter setting for each HPO algorithm.

\subsection{Hardware description}
The numerical experiments were performed using Summit \cite{summit}, a supercomputer at the Oak Ridge Leadership Computing Facility (OLCF) at Oak Ridge National Laboratory. 
Summit has a hybrid architecture; each node contains two IBM POWER9 CPUs and six NVIDIA Volta V100 GPUs all connected together with NVIDIA’s high-speed NVLink. Each node has over half a terabyte of coherent memory (high bandwidth memory + DDR4) addressable by all CPUs and GPUs plus 800 GB of non-volatile RAM that can be used as a burst buffer or as extended memory. To provide a high rate of I/O throughput, the nodes are connected in a non-blocking fat-tree using a dual-rail Mellanox EDR InfiniBand interconnect. 

\subsection{Dataset description}
{ The datasets used are standard benchmark datasets in machine learning, open source and accessible to everyone, and guarantee reproducibility of the results presented. }
The datasets used for the numerical experiments of this section are summarized in Table \ref{tab:datasets}. The dataset \texttt{Eggbox} is artificially constructed by evaluating the function $f(x,y) =  [2+ \cos(x/2) * \cos(y/2)]^5$ across 4,000 points in the domain square $[0,2\pi]^2$ and it is used as a regression problem. The \texttt{Graduate admission} dataset \cite{kaggle} is a regression problem that relates the chances of a student's admission to GRE score, TOEFL score, university rating, and other performance metrics. The \texttt{Computer hardware} dataset \cite{computer_dataset, uci} is a regression problem that describes the relative CPU performance data in terms of its cycle time, memory size, and other hardware properties. The \texttt{Phishing website} dataset \cite{uci} is a classification problem that describes the properties of different websites and classifies them as authentic or fake. The \texttt{CIFAR-10} dataset \cite{cifar10} requires solving a classification problem to classify object images into ten categories. 
The numerical experiments presented in this section are split between the use of MLP and CNN models.
The choice of one type of architecture over the other is dictated by the structure of the dataset used to train the NN models. 
MLP models are used on the \texttt{Eggbox}, \texttt{Graduate admission}, \texttt{Computer hardware}, and \texttt{Phishing website} datasets, whereas CNN models are used for the \texttt{CIFAR-10} dataset. 

\begin{table}[t]
    \centering
    \begin{tabular}{|c|c|c|}
    \hline
    \textbf{Name of dataset} & \textbf{Nb. attributes} & \textbf{Nb. data points} \\
    \hline
         Eggbox &  2 & 4,000 \\ \hline
         Graduate admission &  7 & 400 \\ \hline
         Computer hardware &  9 & 209 \\ \hline
         Phishing websites &  29 & 11,055 \\ \hline
         CIFAR-10 &  - & 60,000 \\ \hline 
    \end{tabular}
    \caption{Description of the datasets. }
    \label{tab:datasets}
\end{table}

\subsection{Training, validation, and test data}
The datasets are split in three components: the training set, the validation set, and the test set. The training set is used to train every instantiated DL model, the validation set is used to select the best performing model at each iteration and the test set is used at the end to measure the predictive power of the NN selected by each HPO algorithm. For the datasets \texttt{Eggbox}, \texttt{Graduate admission}, \texttt{Computer hardware}, and \texttt{Phishing website}, the test set is 10\% of the original dataset, the remaining portion is partitioned into training and validation in the percentage of 90\% and 10\% respectively. For classification problems, a stratified splitting is performed to ensure that the proportion between classes is preserved across training, validation, and test sets. The partitioning between training/validation set and test set for the \texttt{CIFAR-10} dataset is performed as suggested by the online sources where the datasets can be downloaded \ref{fig:cifar10}. 

The optimizer used to train the model is the Adam method \cite{adam} with an initial learning rate of 0.001. 
{We highlight that the number of epochs to train a neural network is different from the number of iterations performed by the hyperparameter optimization algorithm. In fact, the number of epochs is related to the computation needed to perform every single model evaluation. For all HPO methods (GSNNA, TPE, BO), the maximum number of epochs used to train the neural networks is set to be equal to $n$, i.e. the number of samples in the training set for each dataset.
{

The actual number of epochs does not necessarily have to be equal to the number n of points in the dataset. If the training is achieved before n, an early stopping is in place to finish the training. 
If the number of epochs reaches n, that means that the neural network still benefits from the training. Of course, if n is too large, which happens for very large datasets, this may impose unwanted burden on the execution time. But that can be mitigated by trying to find a balance between optimal training and training time. 

}
}

The cost to train a neural network depends on both the size of the dataset, and on the size of the neural network itself. The larger the neural network and the datasets, the longer it takes to train the neural network. The longer time to train a larger neural network on a larger dataset would translate into an increased computational time to perform every single model evaluation, and this would impact the total time to solution for all the HPO algorithms used. 

We also want to point out that the size of a neural network should correlate with the complexity of the relation between inputs and outputs. Having a larger dataset does not necessarily imply needing a larger neural network. 
For example, one may have infinitely many points aligned on a straight line. The dataset is large, but the complexity required for the predictive model to capture the trend is still very low. 

\subsection{Setting of the hyperparameter space}
{\it The hypercube that delimits the hyperparameter search is defined so as to restrict the hyperparameter search within an affordable computational budget}. 
{ Due to the computational budget constraint, we limit the maximum number of layers $L$ to 5.} The number of neurons (or channels) per layer spans from 1 to the highest integer smaller than $\sqrt{n}$, where $n$ is the number of sample points. The choice of $\sqrt{n}$ as the upper bound of the number of neurons per layer is a common practice adopted in DL to avoid overfitting. 
The set of activation functions is made of the sigmoid function (denoted as \texttt{sigmoid} in the Tables), the hyperbolic tangent (\texttt{tanh}), the rectified linear unit function ($\texttt{relu}$) and the exponential linear unit function (\texttt{elu}). The kernel size for CNN architecture spans between 2 and 5. The discrete range for the batch size spans from 10 to the closest integer to $\frac{n}{10}$. Also choosing $\frac{n}{10}$ as maximum size of data batches is a reasonable recommendation adopted by DL practitioners to cap the computational cost of each training iteration. 
The range of search for each hyperparameter is fixed in every HPO algorithm used for the study. Tables \ref{tab:mlp_hyperparameter} and \ref{tab:cnn_hyperparameter} contain a description of the hyperparameters optimized for MLP and CNN architectures with the ranges spanned for each hyperparameter during the optimization. 

\begin{table}[hbt]
    \centering
    \begin{tabular}{|c|c|}
    \hline
    \textbf{Hyperparameter} & \textbf{Search range} \\
    \hline
         Number of hidden layers & \{1,2,3,4,5\} \\ \hline
         Number of neurons per layer & [1,$\sqrt{n}$]  \\ \hline
         nonlinear activation function & \{\texttt{relu}, \texttt{sigmoid}, \texttt{tanh}, \texttt{elu} \} \\ \hline
         batch size & [10, $\frac{n}{10}$] \\ \hline
    \end{tabular}
    \caption{Hyperparameters optimized for MLP architectures. The value $n$ refers to the size of the dataset.}
    \label{tab:mlp_hyperparameter}
\end{table}

\begin{table}[htb]
    \centering
    \begin{tabular}{|c|c|}
    \hline
    \textbf{Hyperparameter} & \textbf{Search range} \\
    \hline
         Number of hidden layers & \{1,2,3,4,5\} \\ \hline
         Number of channels per layer & [1,$\sqrt{n}$] \\ \hline
         Dropout rate & [0,1]\\ \hline
         Pooling & \{1,2\} \\ \hline
         nonlinear activation function & \{\texttt{relu}, \texttt{sigmoid}, \texttt{tanh}, \texttt{elu} \}\\ \hline
         batch size & [10, $\frac{N}{10}$]\\ \hline
    \end{tabular}
    \caption{Hyperparameters optimized for CNN  architectures. The value $n$ refers to the size of the dataset.}
    \label{tab:cnn_hyperparameter}
\end{table}

\subsection{Setting of the hyperparameter search algorithms}
The code to perform GSNNA is implemented in \texttt{python 3.5}, and the NN models are built using \texttt{Keras.io} \cite{keras} which calls \texttt{Tensorflow 2.0} backend. The training of the NN models is performed using the GPUs on Summit by calling \newline \texttt{cudadnn 9.0} for tensor algebra operations. We compare the GSNNA described in this paper with the TPE and BO. 
The version of GSNNA that we implemented performs concurrent model evaluations for the RS at each step with a distributed memory parallelization paradigm that uses \texttt{mpi4py} \cite{mpi4py}. 
The version of TPE and BO used are provided by the \texttt{Ray Tune} library \cite{raytune} through the routines named \texttt{HyperOptSearch} and \texttt{BayesOptSearch} respectively. The version of Ray Tune used is 0.3.1. 
As to \texttt{BayesOptSearch}, the utility function is set to \newline \texttt{utility\_kwargs={"kind": 'ucb'}}, \texttt{"kappa": 2.5}, \texttt{"xi": 0.0}. For both \texttt{HyperOptSearch} and \texttt{BayesOptSearch}, the model selection and evaluations are scheduled using the asynchronous version of HyperBand \cite{Li} called \hfill \newline  \texttt{AsyncHyperBandScheduler}. 
The time attribute for the scheduler is the training iteration and the reward attribute is the validation score of the NN. The validation score is also used as the stopping criterion of the HPO algorithm. 
{ Additional parameters for RayTune’s TPE and BO not mentioned here have been left to default value. Our proposed method, GSNNA, is at its first implementation, whereas the RayTune library used to perform HPO with TPE and BO has underwent multiple stages of implementation optimization. Therefore, our comparison between GSNNA, TPE, and BO does not advantage GSNNA over the other HPO algorithms in terms of implementation.}

\section{Numerical results}
\label{numerical_results}
In this section we present numerical experiments for the five benchmark datasets described above, and we focus on the best suited type of neural network structure for each one of the selected datasets. 
Our numerical experiments compare the performance of GSNNA against BO and TPE in terms of final attainable accuracy of the selected NN architecture and time-to-solution to complete the hyperparameter search. 

{ Numerical tests described in this section focus on weak scaling, meaning that the performance of HPO algorithms is monitored for increased numbers of concurrent model evaluations, with each concurrent model evaluation mapped to a separate MPI process and a separate GPU to train, and the predictive performance of the model is assessed. 
Strong scaling tests are not included in the discussion for the following reasons. 
For applications such as the ones considered in this paper, strong scaling requires fixing the number of concurrent model evaluations and progressively increase the computational resources made available for each model evaluation. In our methodology, there is a one-to-one mapping between concurrent model evaluations and GPUs. When the total number of GPUs is less than the concurrent models, the strong scaling boils down to the scaling of the job scheduler, which is outside the scope of this work. When the total number of GPUs is more than the concurrent models, this would translate to using multiple GPUs to perform a single model evaluation instead of using one GPUs as currently done in the work. In the deep learning community, this approach is known as model parallelization.
Model parallelization would accelerate the model evaluations and it would equally apply to all the three methods TPE, BO, and GSNNA. However, model parallelization would not accelerate the execution of the hyperparameter optimization algorithms themselves. Therefore, the comparison of TPE, BO, and GSNNA would not differ from the ones presented in this paper in relative terms. 
Moreover, the small size of the neural networks and the small size of the benchmark datasets used in this work does not justify model parallelization; strong scaling would only bring marginal benefits on the acceleration of model evaluations.}

\subsection{Comparison for predictive performance and computational time}
\label{section_comparison}
The first set of numerical experiments compares the predictive power of the GSNNA with TPE and BO. The metric used to quantify the predictive performance of a NN for regression problems is the $R^2$ score defined as 
\begin{equation}
R^2 = 1 - \frac{\sum_{i=1}^m (y_i - \hat{y}_i)^2}{\sum_{i=1}^m (y_i - \overline{y}_i)^2}
\end{equation}
where $y_i$ are the observations for $m$ data points in the test set, $\hat{y}_i$ are the predictions obtained with the DL model over the test set and $\bar{y}_i$ is the sample mean of the data points over the test set. The metric used to quantify the predictive performance of a NN used for classification problems is the $F1$ score defined as 
\begin{equation}
F1 = 2\frac{PPV \cdot TPR}{PPV + TPR},    
\end{equation}
where $PPV = \frac{\text{true positives}}{\text{true positives}+\text{false negatives}}$ is the \textit{precision} or \textit{positive predicted value} and $TPR = \frac{\text{true positives}}{\text{positives}}$ is the \textit{sensitivity}, \textit{recall}, \textit{hit rate}, or \textit{true positive rate}.   

For the datasets that require the use of MLP architectures, the number of concurrent model evaluations per iteration is set to 10, 25, 50, 75, and 100 for all the three HPO algorithms. For the \texttt{CIFAR-10} dataset that requires the use of CNN architectures, the number of concurrent model evaluations per iteration is set to 150, 300, 450, and 600 to cope with a larger number of hyperparameters to tune. { }
The maximum number of iterations is set to 5 for all the three HPO algorithms and the stopping criterion imposes a threshold on the $R^2$ score and $F1$ score equal to 0.99. 

{ To guarantee a fair comparison between the different HPO algorithms, the implementations of the three HPO algorithms make use of the same number of concurrent model evaluations, and each implementation of the HPO algorithms maps every concurrent model evaluation to a separate GPU. However, the complexity of (and thus the cost to train) each model per iteration varies according to the specific architectures that the HPO algorithms select at each iteration. Since different HPO algorithms select different architectures to construct and evaluate, this can lead to different computational times. Because Summit has six GPUs per compute node, the total number of Summit nodes used in a numerical experiment is equal to the least integer greater than or equal to the concurrent model evaluations divided by 6. }

\begin{figure}[ht]
   \centering
\begin{tabular}{c}
\includegraphics[width=0.98\columnwidth]{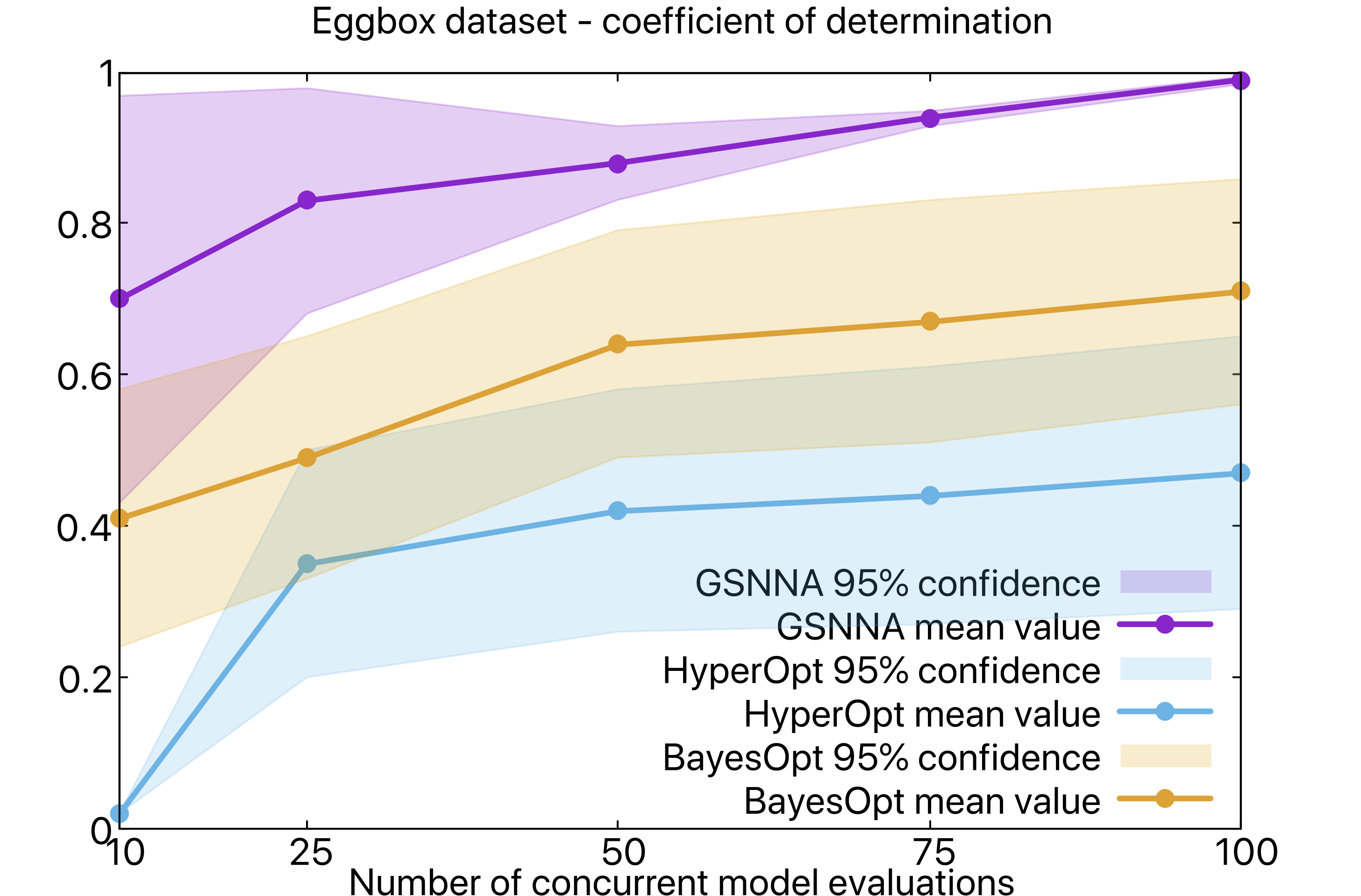} \\ \\
\includegraphics[width=0.98\columnwidth]{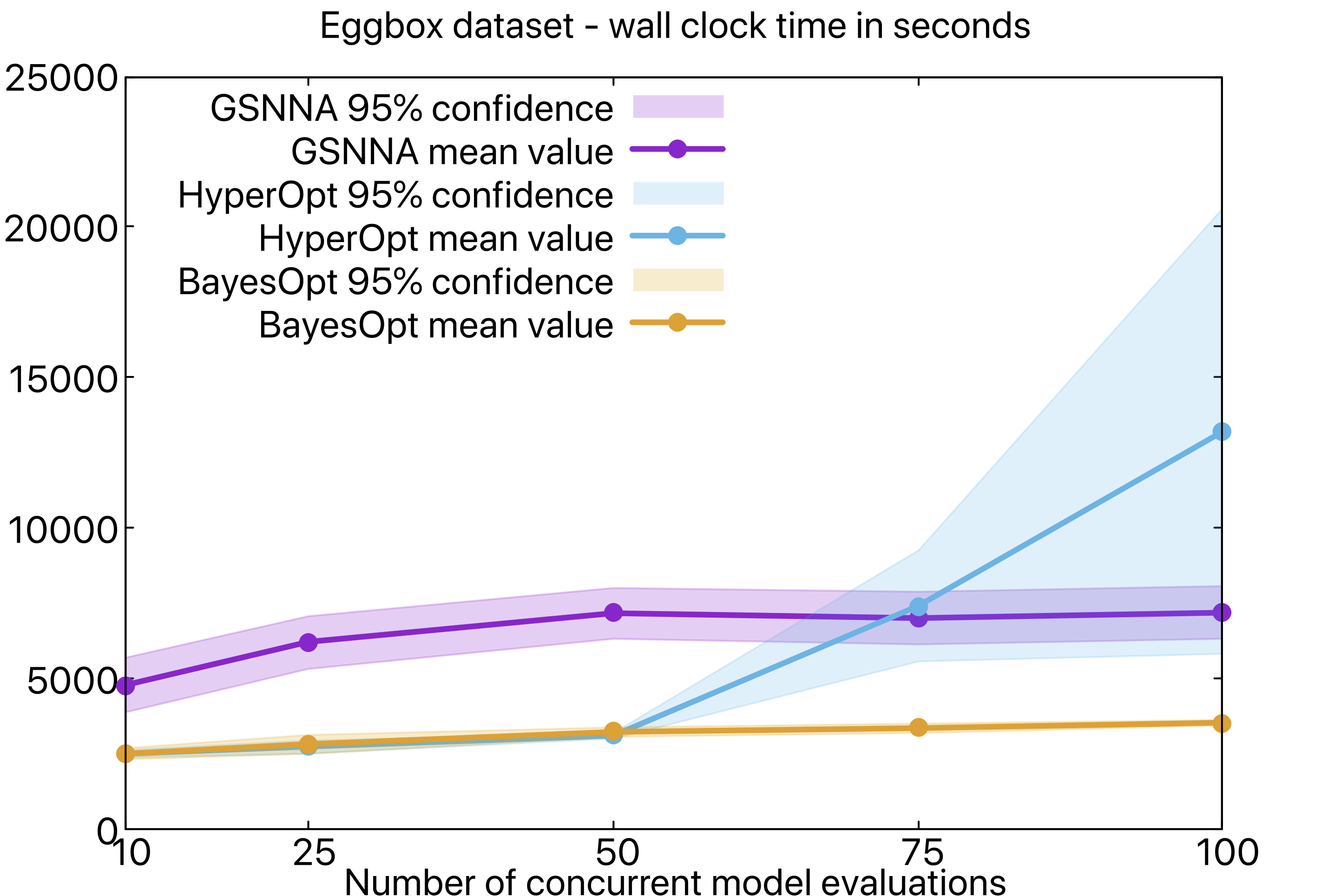}
\end{tabular}
    \caption{\texttt{Eggbox} dataset. Comparison between Greedy search, HyperOptSearch and BayesOptSearch for test cases with MLP architectures. The graph at the top shows the performance obtained by the model selected by the hyperparameter search on the test set. The graph at the bottom shows a comparison of the computational times. }
    \label{fig:eggbox}
\end{figure}

\begin{figure}[ht]
   \centering
\begin{tabular}{c}
\hspace{0.3cm}\includegraphics[width=0.95\columnwidth]{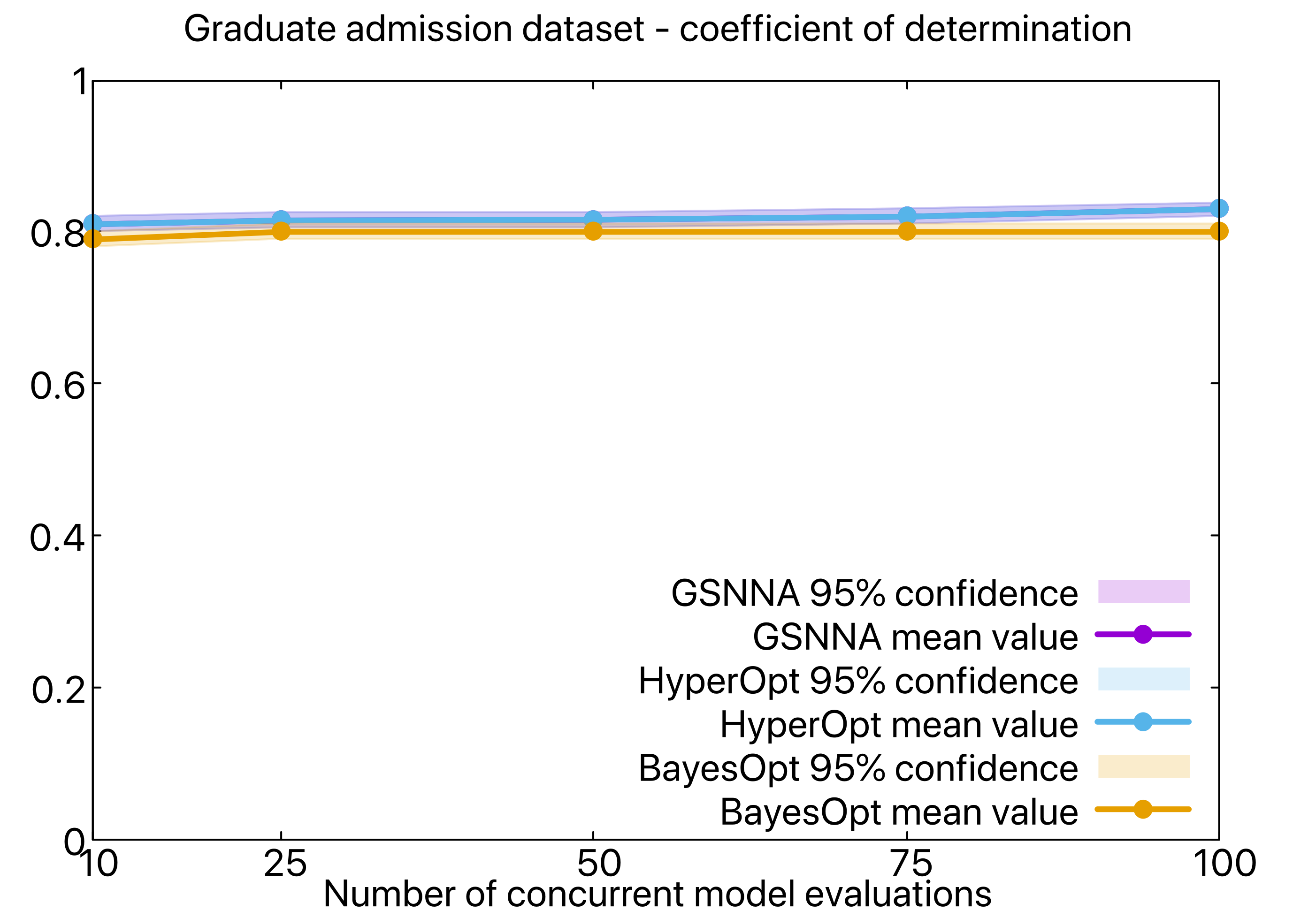}\\ \\
\includegraphics[width=0.98\columnwidth]{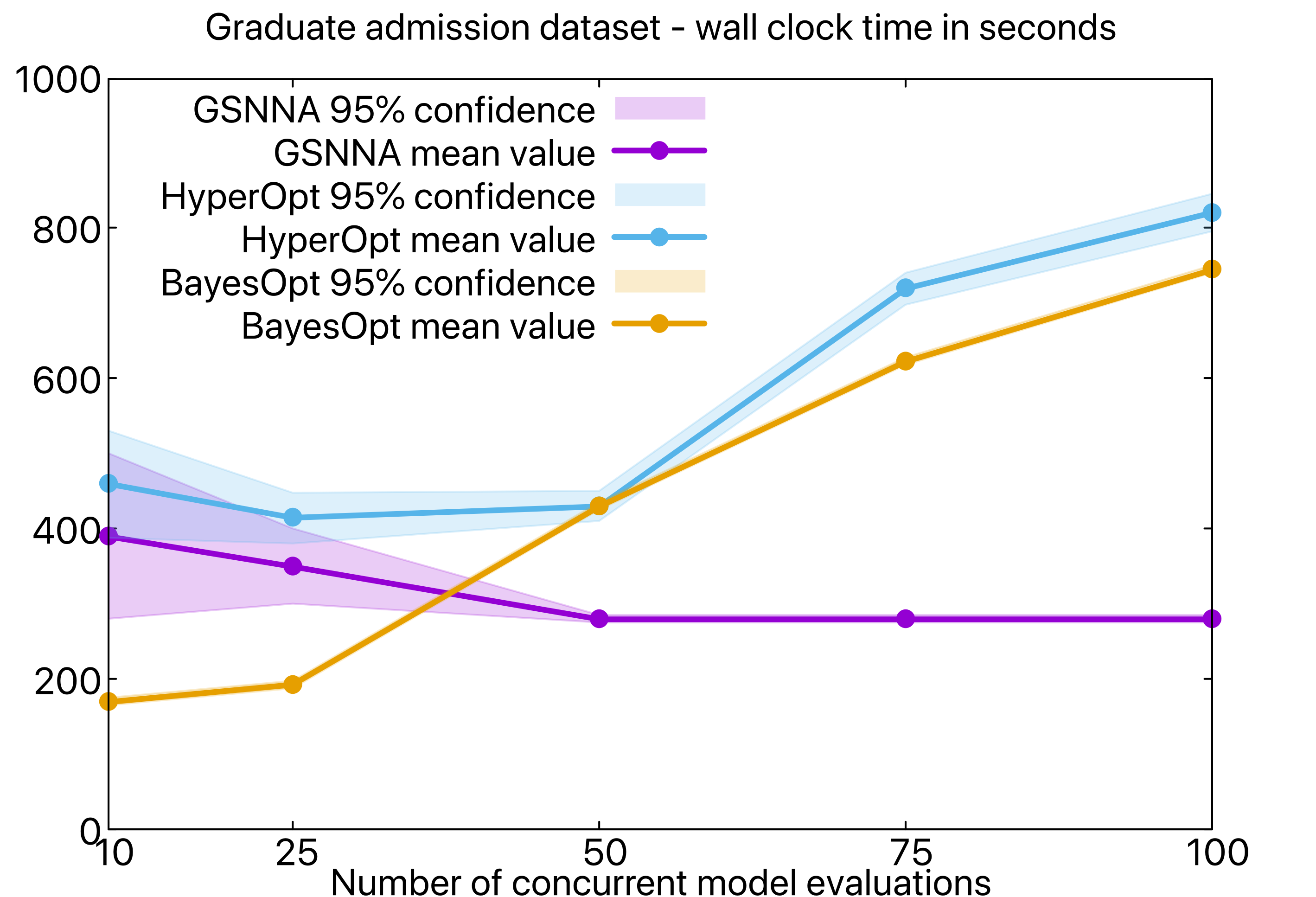}
\end{tabular}
    \caption{\texttt{Graduate admission} dataset. Comparison between Greedy search, HyperOptSearch and BayesOptSearch for test cases with MLP architectures. The graph at the top shows the performance obtained by the model selected by the hyperparameter search on the test set. The graph at the bottom shows a comparison of the computational times. }
    \label{fig:graduate} 
\end{figure}

\begin{figure}[ht]
   \centering
\begin{tabular}{c}
\includegraphics[width=0.945\columnwidth]{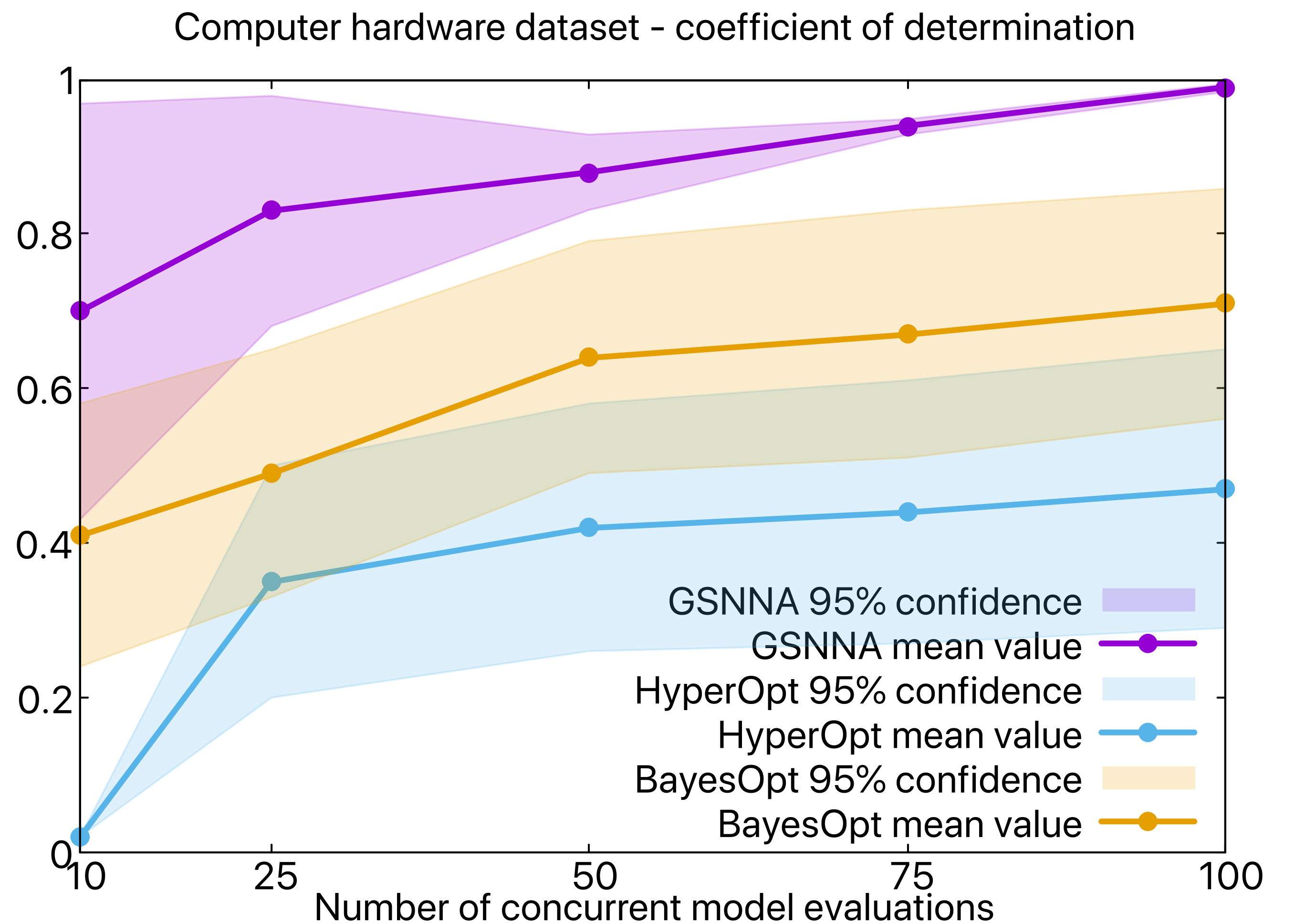}\\ \\
\includegraphics[width=1.0\columnwidth]{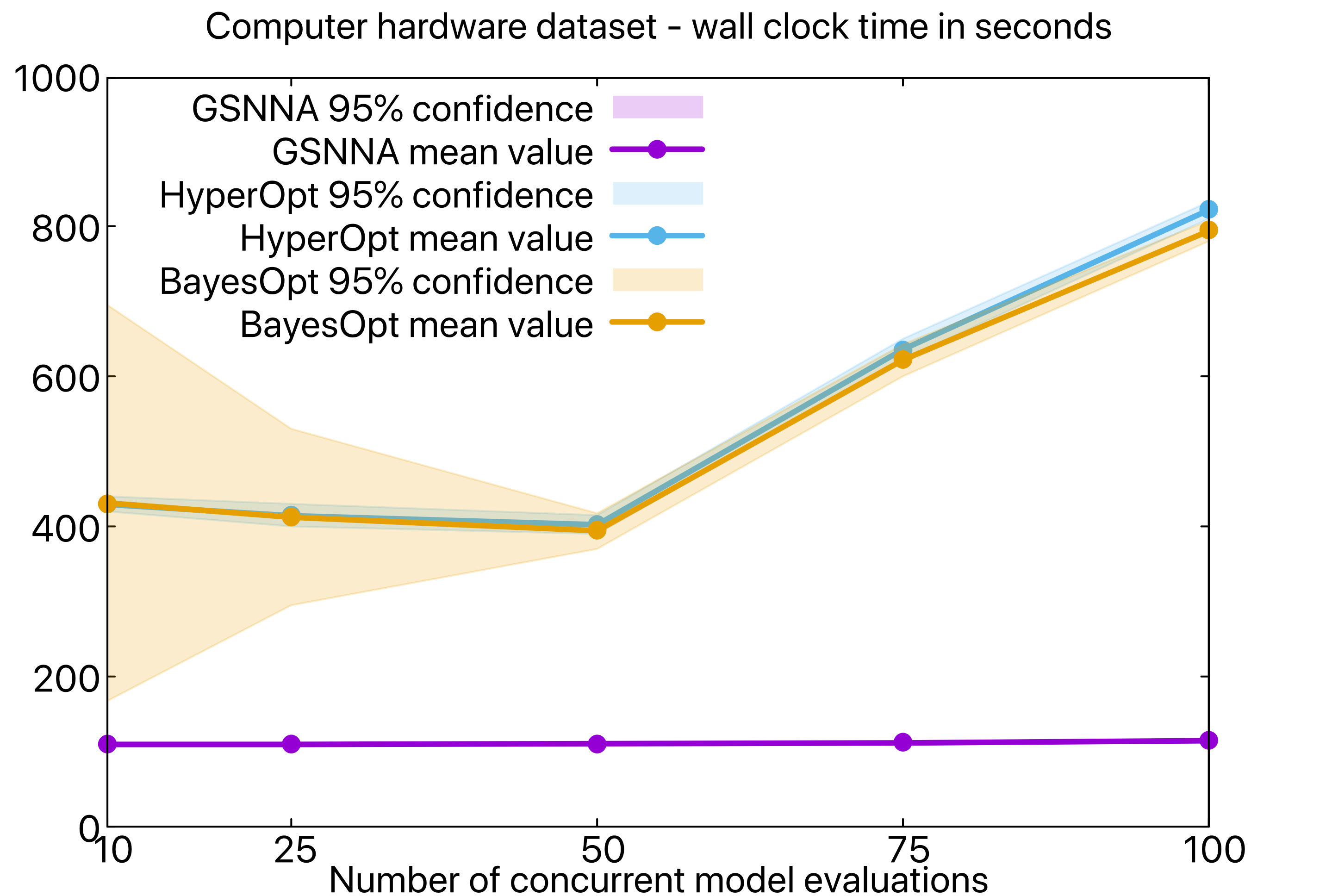}
\end{tabular}
    \caption{\texttt{Computer hardware} dataset. Comparison between Greedy search, HyperOptSearch and BayesOptSearch for test cases with MLP architectures. The graph at the top shows the performance obtained by the model selected by the hyperparameter search on the test set. The graph at the bottom shows a comparison of the computational times. }
    \label{fig:cpu} 
\end{figure}

\begin{figure}[ht]
   \centering
\begin{tabular}{c}
\hspace{0.3cm}\includegraphics[width=0.95\columnwidth]{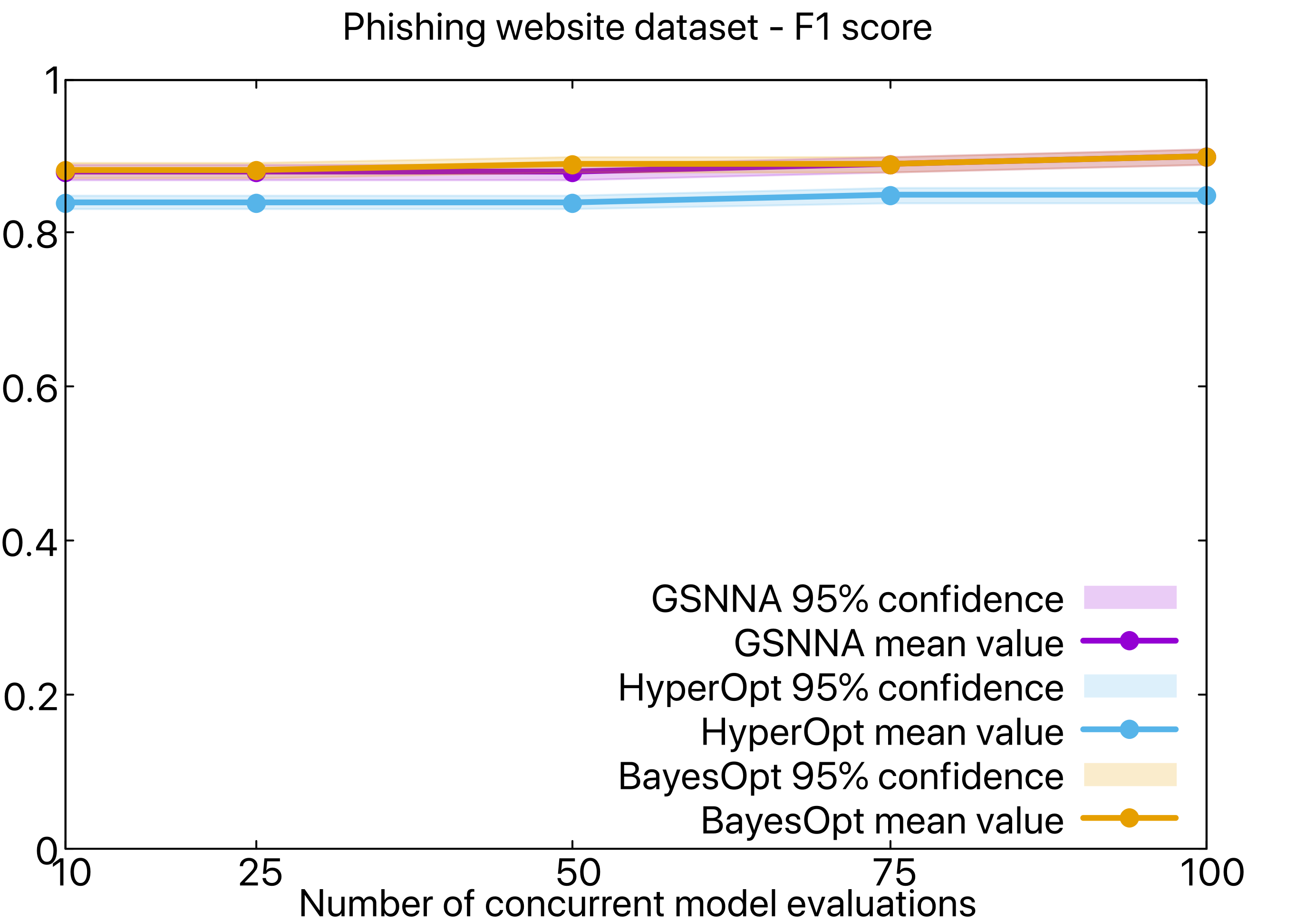} \\ \\
\includegraphics[width=0.98\columnwidth]{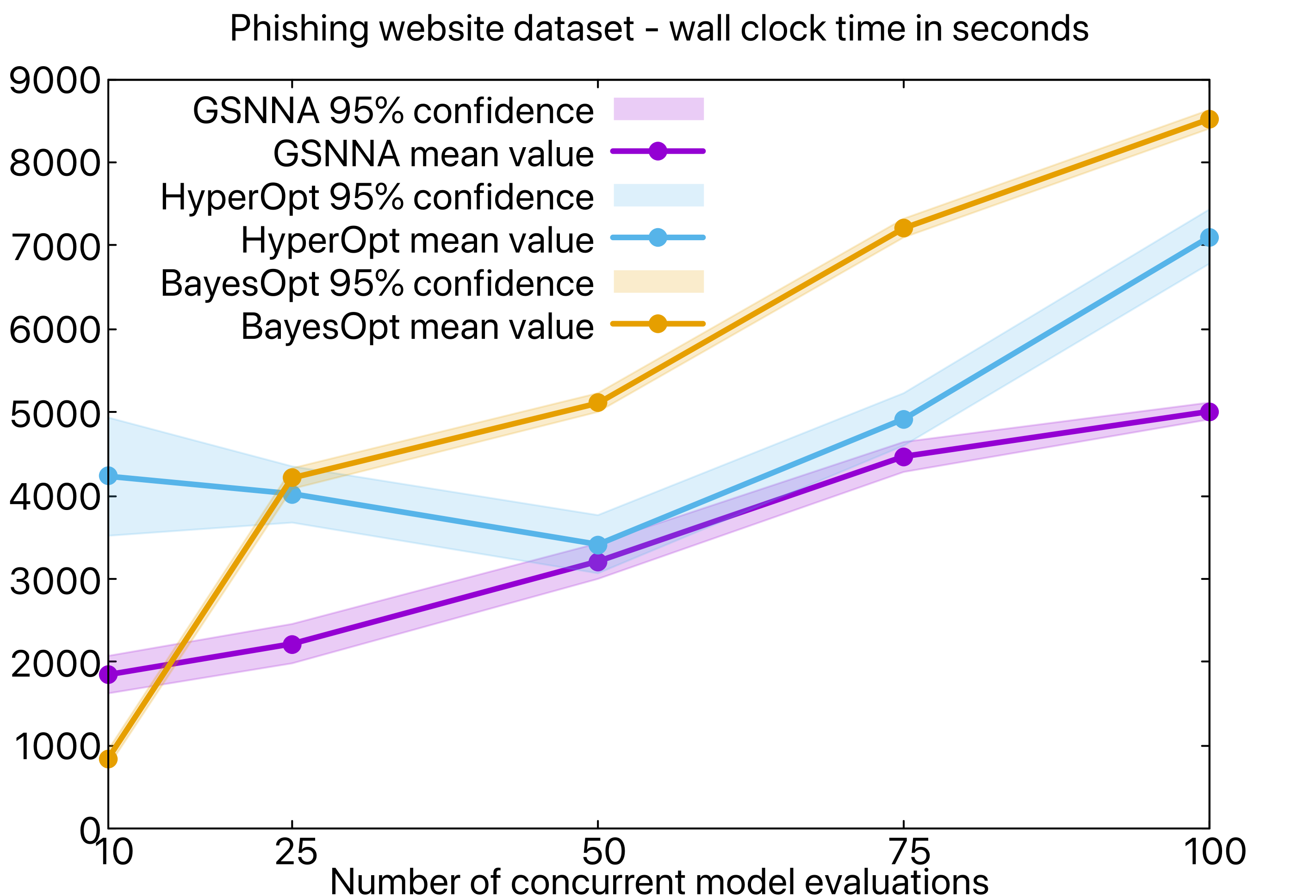}  \\
\end{tabular}
    \caption{\texttt{Phishing} dataset. Comparison between Greedy search, HyperOptSearch and BayesOptSearch for test cases with MLP architectures. The graph at the top shows the performance obtained by the model selected by the hyperparameter search on the test set. The graph at the bottom shows a comparison of the computational times. }
    \label{fig:phishing} 
\end{figure}

Figures \ref{fig:eggbox}, \ref{fig:graduate}, \ref{fig:cpu}, and \ref{fig:phishing} correspond to the test cases with MLP models. In these figures, the figures on top show the scores obtained on the test set of the selected MLP model, and the figures at the bottom show the time-to-solution in wall clock seconds. The performance is reported for each hyperparameter search algorithm, averaging over 10 runs with 95\% confidence intervals both for the mean value of the predictive performance and for the mean value of the time-to-solution. 

The experiments with the \texttt{Eggbox} dataset exhibit better results for GSNNA with respect to TPE and BO in terms of predictive power achieved by the selected NN. Moreover, we notice that the confidence band for GSNNA narrows as the number of concurrent evaluations increases. This happens because the inference on the attainable predictive performance becomes more accurate with a higher number of random samples for the stratified RS. A different trend is shown for the confidence band of TPE and BO. In this case, the confidence band does not become narrower by increasing the number of concurrent model evaluations. This highlights the benefit of using a stratified RS in GSNNA: the uncertainty of the random optimization is bounded by reducing the dimensionality of the search space. 

In terms of scalability, we notice that GSNNA has a flat weak scaling curve, whereas BO and TPE significantly increase the computational time-to-solution with an increased number of concurrent model evaluations. Although BO and TPE finish in less time than GSNNA for 10 and 50 model evaluations, the final attained accuracy is significantly lower than the one obtained with GSNNA. This indicates that GSNNA better explores the hyperparameter space.

Similar results in terms of final attainable accuracy and scalability have been obtained for \texttt{Graduate admission}, \\ \texttt{Computer hardware} and \\ \texttt{Phishing websites} datasets. 
Although different values for the tuning parameter of \texttt{BayesOptSearch} have been tested on the datasets considered in this paper, we noticed that the performance of \texttt{BayesOptSearch} on these datasets did not significantly change.
{ We also noticed that for the \texttt{graduate admission} dataset and the \texttt{phishing} dataset, some HPO algorithms reduce the total time of the search for an increased number of concurrent model evaluations, and this goes against an intuitive reasoning. To better understand this phenomenon, we note that the number of concurrent models impacts the computational time in two ways: a higher number of concurrent model evaluations makes it likely to identify a network that attains a desired accuracy faster, but it also needs more time to coordinate the model evaluations between each other. Whether one of these two factors prevails over the other can results in either a reduction or an increase in the total computational time. }

\begin{figure}[htbp]
   \centering
\begin{tabular}{c}
\includegraphics[width=9.cm]{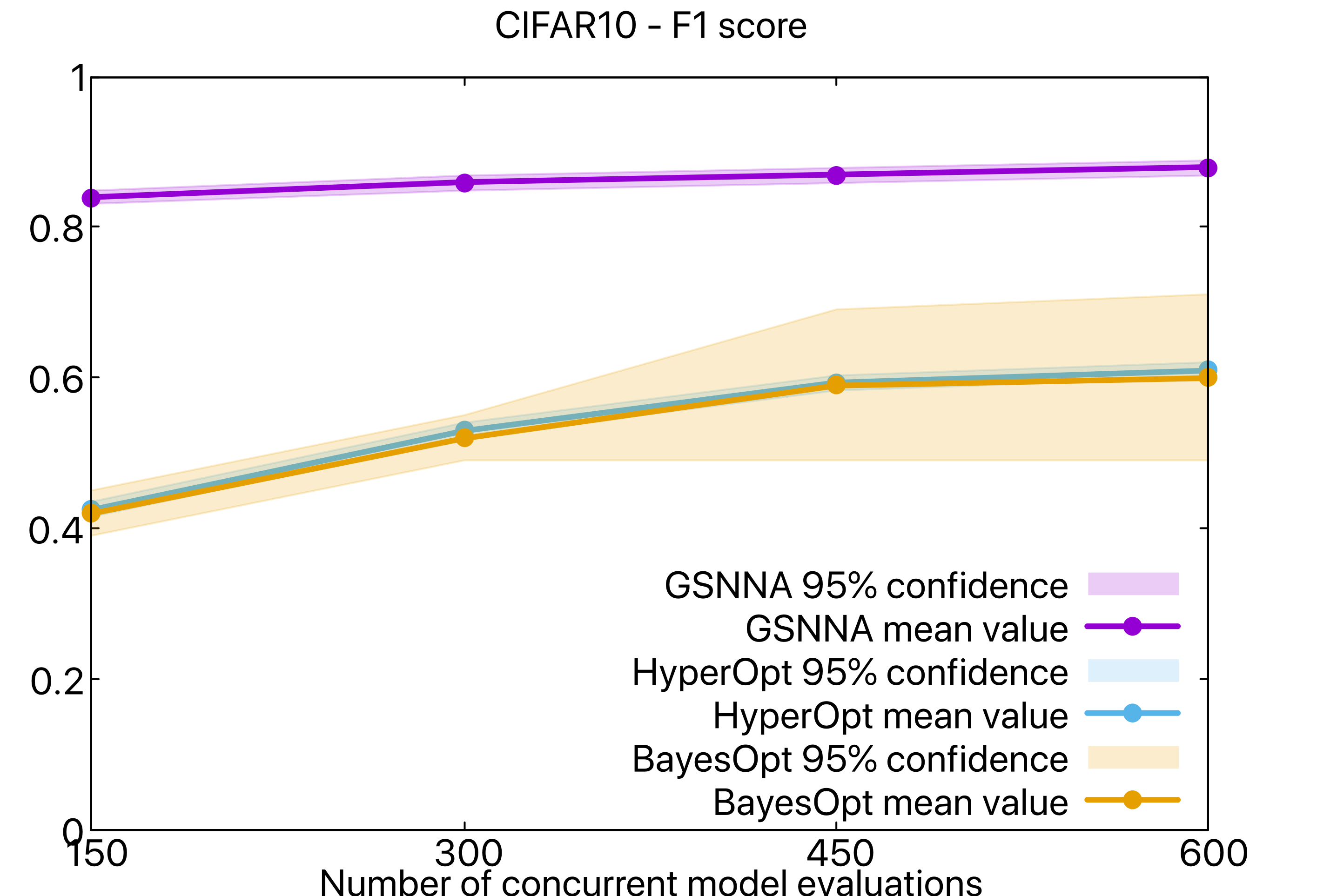}\\ \\ 
\hspace{-0.2cm}\includegraphics[width=8.9cm]{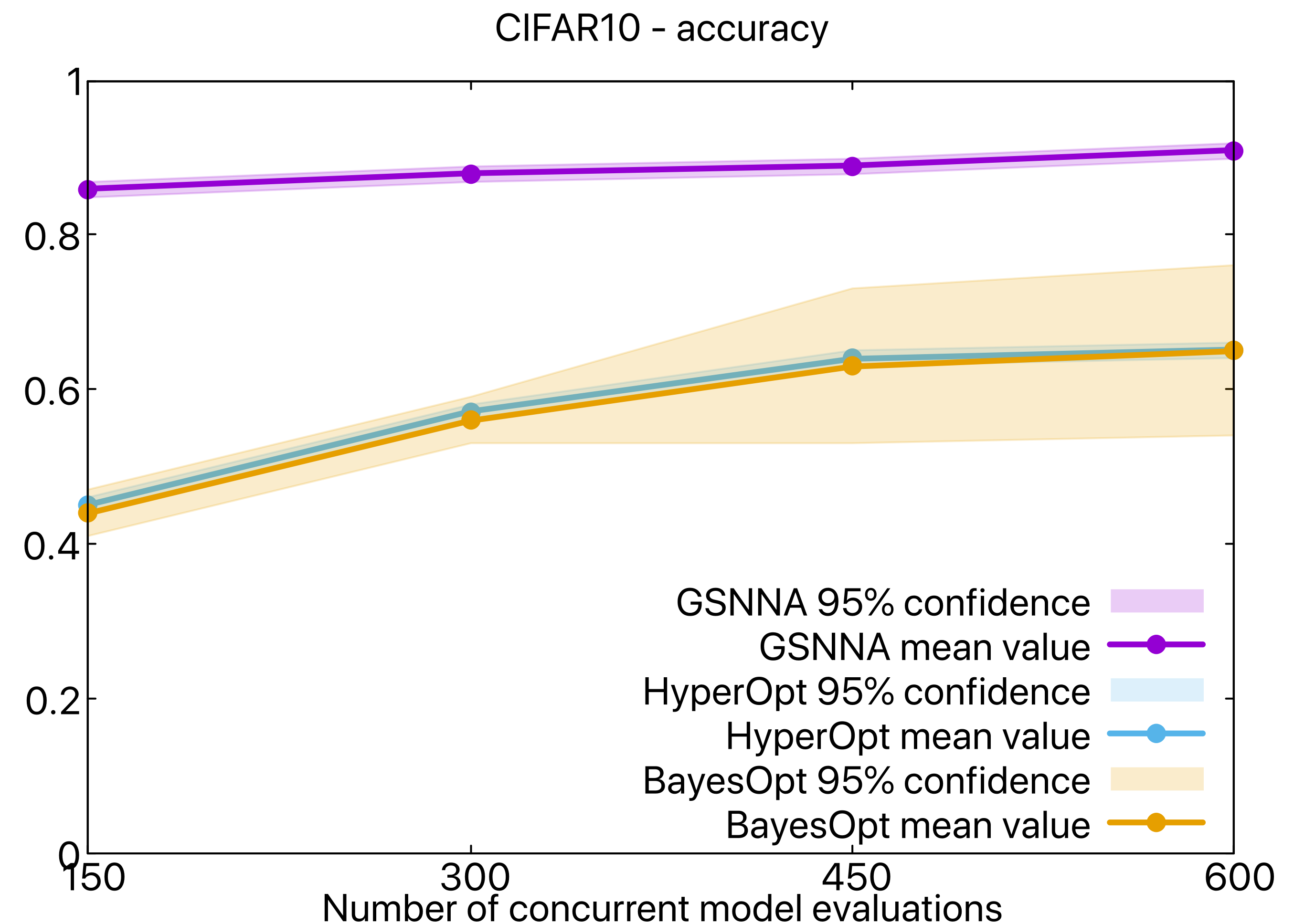} \\ \\
\hspace{-0.4cm}\includegraphics[width=9.7cm]{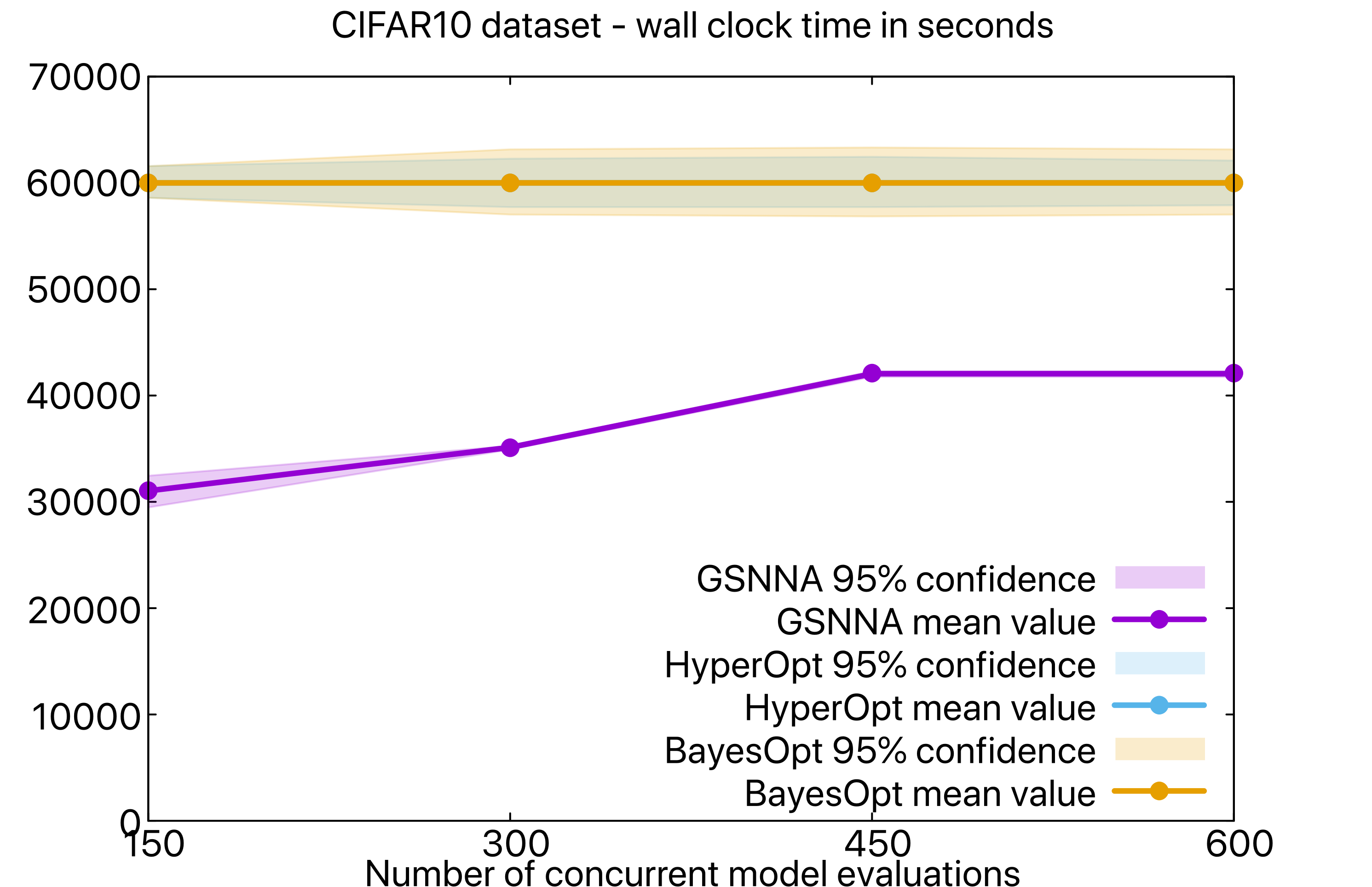}
\end{tabular}
    \caption{Comparison between GSNNA, HyperOptSearch and BayesOptSearch for test cases with CNN architectures. The comparison is performed for the CIFAR10 dataset. The graph on top shows the performance obtained by the model selected by the hyperparameter search on the test set in terms of F1 score. The graph in the center shows the performance obtained by the model selected by the hyperparameter search on the test set in terms of accuracy. The graph at the bottom shows the computational time.}
    \label{fig:cifar10} 
\end{figure}

 The results for the \texttt{CIFAR-10} dataset using CNN in Figure \ref{fig:cifar10} show that GSNNA outperforms both TPE and BO algorithms in terms of best attainable predictive performance and computational time. { The F1-score is more appropriate than the accuracy (percentage of data points correctly classified) to measure the predictive performance of neural networks for classification purposes in case of class imbalance \cite{tharwat}. However, the accuracy is still the mostly used metric to report the predictive performance of a model on some benchmark dataset such as \texttt{CIFAR-10}. In order to facilitate the comparison with other results published in the literature, we also report the accuracy for \texttt{CIFAR-10}.}
 
 { Comparing the architecture selected by GSNNA with state-of-the-art architectures customized for \texttt{CIFAR-10} \cite{cifar10_list}, we see that the predictive performance of our architecture has a test error of about 9\%, whereas customized architectures currently provide error below 0.1\%. In view of this gap between the performance we obtained on \texttt{CIFAR-10} with respect to other results published in the literature, we emphasize that the goal of our research is to build an automatic selection of hyperparameters that is as agnostic as possible about the specifics of the dataset at hand. This makes the hyperparameter search more challenging, and the attainable accuracy is generally lower than the one obtained with customized approaches.
 Recent results obtained by other researchers \cite{wu} show a test error around 12\% when a Bayesian approach is used to optimize the architecture of a neural network for the \texttt{CIFAR-10} dataset, and this is in line with the results we present here.}

\subsection{Sensitivity of GSNNA with respect to the number of concurrent model evaluations}
\begin{figure}[htbp]
   \centering
\begin{tabular}{c}
\includegraphics[width=9cm]{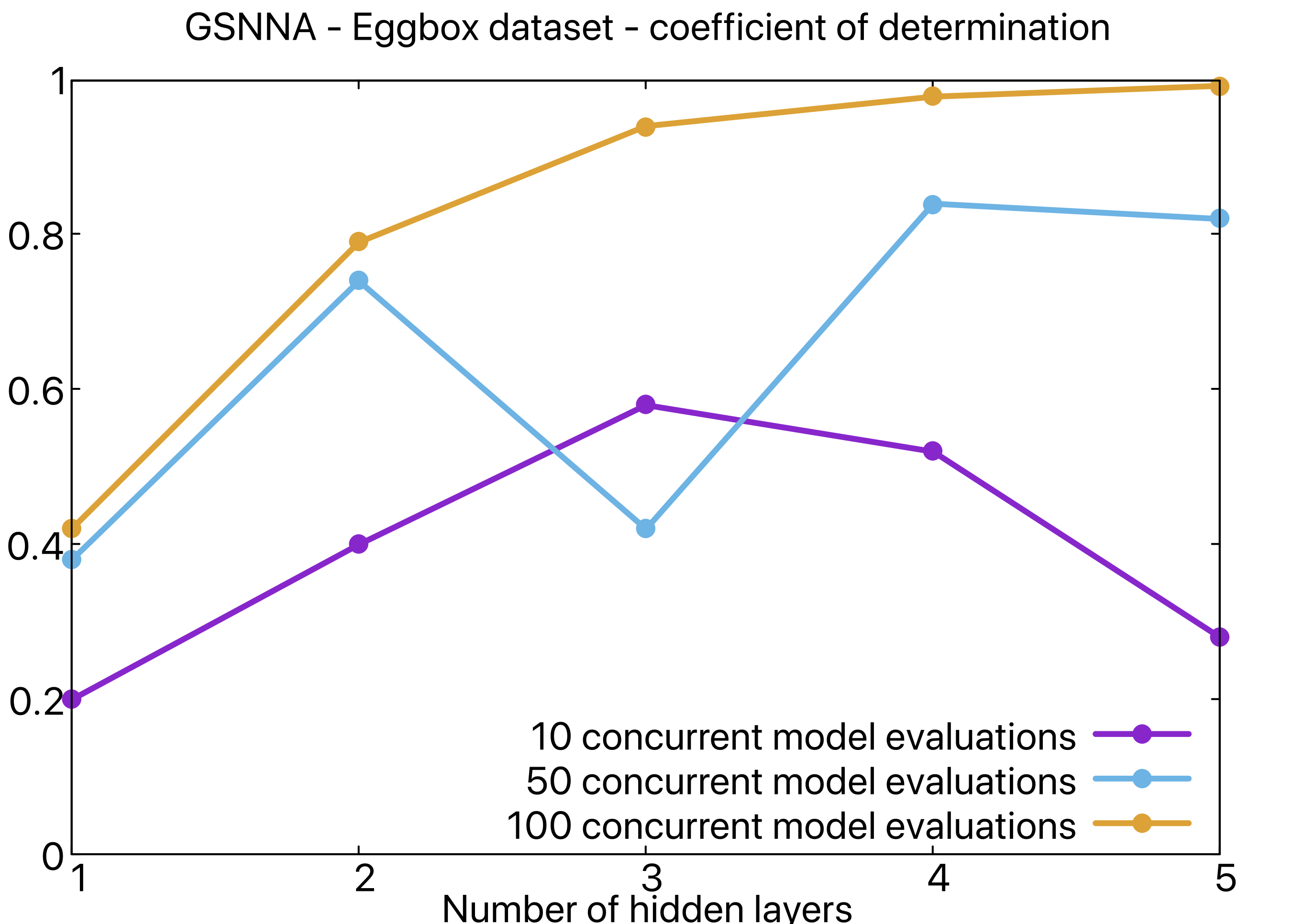} \\ \\
\includegraphics[width=9cm]{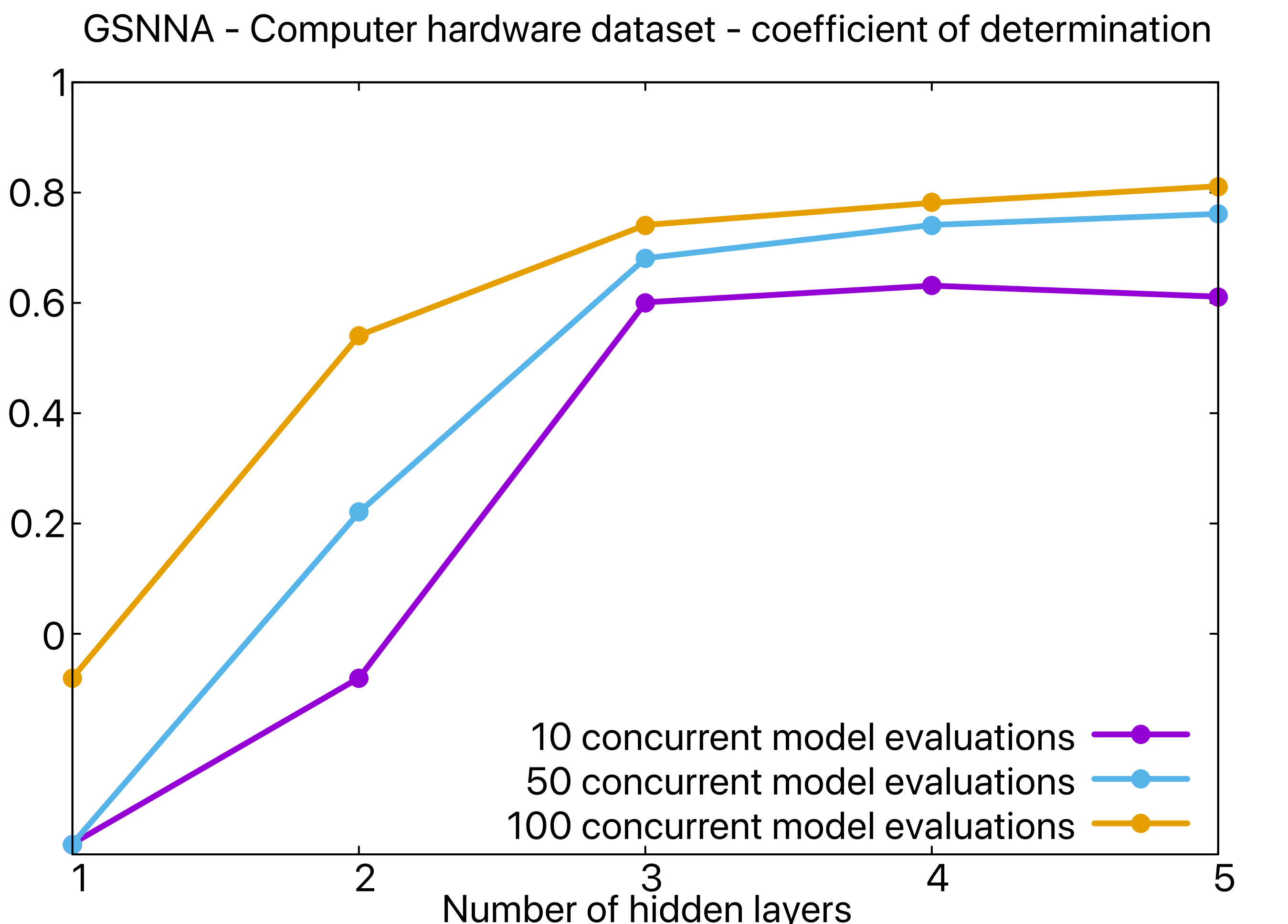}
\end{tabular}
    \caption{Greedy Search for Neural Network Architecture (GSNNA). Coefficient of determination expressed in terms of the number of hidden layers for \texttt{Eggbox}, \texttt{Computer hardware} datasets using 10, 50, and 100 concurrent model evaluations. Results are shown for a single run.}
    \label{fig:fig2} 
\end{figure}

In Figure \ref{fig:fig2} we show the performance obtained with GSNNA on the \texttt{Eggbox} dataset and the \texttt{Computer hardware} dataset as a function of the number of hidden layers for different numbers of concurrent model evaluations (10, 50, and 100). For both experiments it is clear that the use of a small number of concurrent model evaluations leads to significant fluctuations in the score, as the stratified RS does not explore enough architectures for a fixed number of hidden layers. A progressive increase in the concurrent model evaluations leads to a better inference. This happens because an exhaustive exploration of the stratified hyperparameter space reduces the uncertainty in the attainable best performance of the model.
Moreover, a sufficient exploration of the stratified hyperparameter space enables us to highlight the dependence between the maximum attainable performance of the NN and the total number of hidden layers. Indeed, the examples displayed in Figure \ref{fig:fig2} confirm that nonlinear input-output relations can benefit from a higher number of hidden layers. 

\begin{figure}[htbp]
   \centering
\begin{tabular}{c}
\includegraphics[width=9cm]{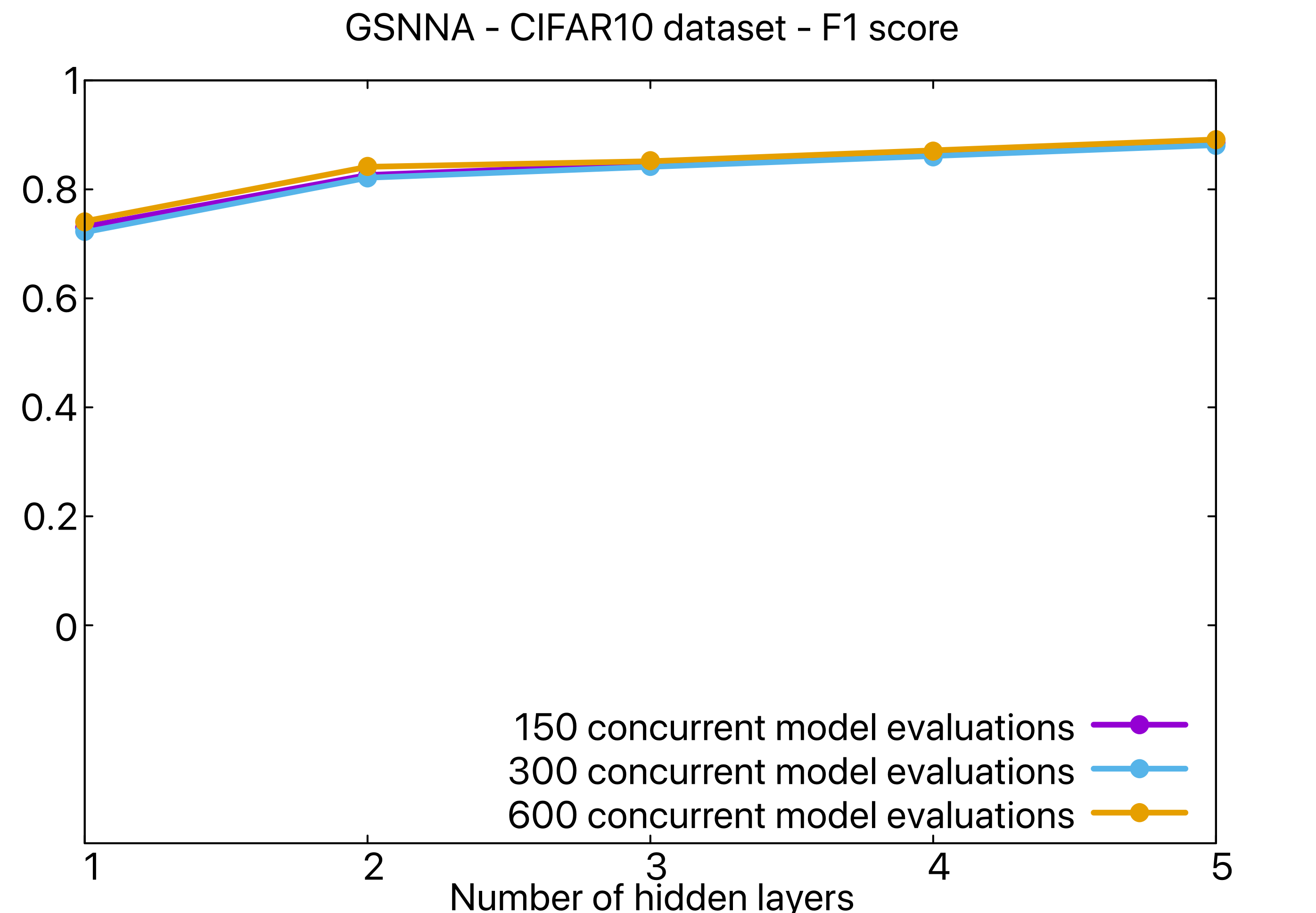}
\end{tabular}
    \caption{Greedy Search for Neural Network Architecture (GSNNA). Coefficient of determination expressed in terms of the number of hidden layers for \texttt{CIFAR10} dataset using 150, 300, and 600 concurrent model evaluations. Results are shown for a single run.}
    \label{fig:cifar10_layers} 
\end{figure}

In Figure \ref{fig:cifar10_layers} we present a similar analysis using CNN for the \texttt{CIFAR-10} dataset. In this case, the number of concurrent model evaluations considered is 150, 300, and 600. {The scalability tests for the \texttt{CIFAR-10} dataset use a higher number of concurrent model evaluations with respect to the previous datasets because there are more architectural hyperparameters to tune in CNN than in MLP models, as described also by a comparison between Tables \ref{tab:mlp_hyperparameter} and \ref{tab:cnn_hyperparameter}}. Different from the previous numerical examples, increasing the number of concurrent model evaluations does not benefit the identification of a better performing architecture for the \texttt{CIFAR-10} dataset, but a progressive increase of the number of hidden layers still leads to a progressive gain in attainable accuracy.

\section{Concluding remarks and future developments}
\label{conclusions}
GSNNA aims to determine in a scalable fashion, and within a given computational budget, the NN with minimal number of layers that performs at least as well as NN models of the same structure identified by other hyperparameter search algorithms.
The algorithm adopts a greedy technique on the number of hidden layers, which can benefit the reduction of computational time and cost to perform the hyperparameter search. This makes the algorithm not only appealing, but sometimes strongly compelling when computational and memory resources are limited, { or when DL driven decisions have to be performed in a timely manner}. 
The recycling of hidden layer configurations disregards an exponential number of architectures in the hyperparameter space. However, having a smaller search space makes the optimization a much more tractable problem with a significant reduction in computational complexity. Moreover, our numerical results show that this does not compromise the final attainable accuracy of the model selected by the optimization procedure. 

{

CIFAR-10 is the largest tested dataset, with 60000 images at 32x32 resolution. ImageNet or the Open Images Dataset have more than a million images and are commonly evaluated at 256x256. At the same efficiency, this could take 1000x more time, and CIFAR-10 already takes about 8 hours. This is a limitation to the applicability of the method. However, the proposed research aims at improving scalability of hyperparameters search algorithms with a {\it constrained computational budget}.  
Therefore, while the method is illustrated on modest-size datasets and neural networks, it has promise for implementations on larger datasets and correspondingly larger neural networks under the same computational budget constraints. } 

For future developments we aim to extend the study to different types of architectures other than multilayer perceptrons and CNN, such as residual neural networks (ResNet), recurrent neural networks (RNN) and long short-term memory neural networks (LSTM). We will also use GSNNA for specific problems by selecting customized attributes other than the score for the HPO, and we will conduct an uncertainty quantification analysis to estimate the sensitivity of the inference on the hyperparameters with respect to the dimension of the hyperparameter space and the number of concurrent model evaluations.

\section*{Acknowledgements}

Massimiliano Lupo Pasini thanks Dr. Vladimir Protopopescu for his valuable feedback in the preparation of this manuscript { and three anonymous reviewers for their very useful comments and suggestions}.

This work is supported in part by the Office of Science of the US Department of Energy (DOE) and by the LDRD Program of Oak Ridge National Laboratory. This work used resources of the Oak Ridge Leadership Computing Facility \\ (OLCF), which is a DOE Office of Science User Facility supported under Contract DE-AC05-00OR22725. Y. W. Li was partly supported by the LDRD Program of Los Alamos National Laboratory (LANL) under project number 20190005DR. LANL is operated by Triad National Security, LLC, for the National Nuclear Security Administration of U.S. Department of Energy (Contract No. 89233218CNA000001). This document number is LA-UR-21-20936.

\FloatBarrier

\section*{References}
\bibliographystyle{unsrt}
\bibliography{sample}

\end{document}